\documentclass[11pt,a4paper,twocolumn]{article}

\usepackage[utf8]{inputenc}
\usepackage{times}
\usepackage[colorlinks=false, pdfborder={0 0 0}]{hyperref}
\usepackage{breakurl}
\usepackage[rightcaption]{sidecap}
\usepackage{stfloats}

\usepackage{subeqn}
\usepackage{listings}

\usepackage{enumitem}

\usepackage[font=small,margin=1ex,labelfont=bf]{caption}

\usepackage{subfigure}

\usepackage{calc}
\usepackage[left=2.0cm,right=2.0cm,top=3.0cm,bottom=3.0cm]{geometry}

\usepackage{parskip}
\usepackage{graphicx}

\usepackage{fancyhdr}
\fancypagestyle{firstpage}{
  \fancyhead{}
  \cfoot{\footnotesize{\em Technical Report No.~2014-01, Hochschule Niederrhein, Fachbereich Elektrotechnik \& Informatik (2014)}}
  
}

\title{\vspace*{-10mm}Soft Thresholding for Visual Image Enhancement}
\author{
Christoph Dalitz\\
Institut f\"ur Mustererkennung\\
Hochschule Niederrhein\\
Reinarzstr. 49, 47805 Krefeld\\
{\tt christoph.dalitz{@}hsnr.de}
}
\date{}

\begin{document}

\renewcommand{\labelenumi}{\arabic{enumi})}

\twocolumn[
  \begin{@twocolumnfalse}
    \maketitle
\begin{abstract}
Thresholding converts a greyscale image into a binary image, and is thus often a necessary segmentation step in image processing. For a human viewer however, thresholding usually has a negative impact on the legibility of document images. This report describes a simple method for ``smearing out'' the threshold and transforming the greyscale image into a different greyscale image. The method is similar to fuzzy thresholding, but is discussed here in the simpler context of greyscale transformations and, unlike fuzzy thresholding, it is independent from the method for finding the threshold. A simple formula is presented for automatically determining the width of the threshold spread. The method can be used, e.g., for enhancing images for the presentation in online facsimile repositories.
\end{abstract}
\vspace*{2ex}

  \end{@twocolumnfalse}
  ]

\thispagestyle{firstpage}

\section{Introduction}
Thresholding can be considered as a special case of image segmentation: it partitions the image pixels of a greyscale image into foreground (typically ``black'') and background (``white'') pixels, thereby transforming the greyscale image into a binary image. As it is both an essential and a possibly difficult preprocessing step in many image processing systems, in particular for document image recognition, many different thresholding techniques have been proposed in the literature \cite{pal93} \cite{sezgin04}. The thresholding algorithm itself is very simple: let $f(x,y)$ be the grey value of the image at pixel position $(x,y)$; then thresholding with threshold $t$ transforms this image into a binary image $\tilde{f}(x,y)$ as follows:
\begin{equation}
\label{eq:thresholding}
\tilde{f}(x,y) = \left\{\begin{array}{l} 1\;\mbox{ if }f(x,y)\leq t \\
0\;\mbox{ if }f(x,y)> t \end{array}\right.
\end{equation}
When the threshold is constant over the entire image, the thresholding is called {\em global}. When it depends on the position, i.e. $t=t(x,y)$, the thresholding is called {\em local}. The different thresholding algorithms vary in their rules for determining the threshold $t(x,y)$. An often deployed algorithm for global thresholding is Otsu's method \cite{otsu79}.

\begin{figure}[!t]
  \begin{center}
  \subfigure[Greyscale image]{
    \includegraphics[width=0.3\textwidth]{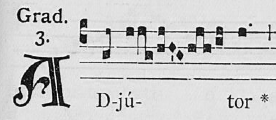}
    \label{fig:thresholding:a}
  }
  \subfigure[Otsu thresholding]{
    \includegraphics[width=0.3\textwidth]{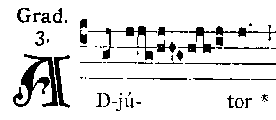}
    \label{fig:thresholding:b}
  }
  \subfigure[Soft thresholding]{
    \includegraphics[width=0.3\textwidth]{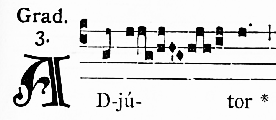}
    \label{fig:thresholding:c}
  }
  \end{center}
  \caption{\label{fig:thresholding}A Greyscale image binarized with Otsu's method and transformed with soft thresholding using the same threshold (image detail from ``Graduale Romanum'', Tournai, 1910).}
\end{figure}

As can be seen in Fig.~\ref{fig:thresholding}, converting a greyscale image to a binary image has the effect that object borders that look smooth in the greyscale image become ragged in the binary image. This negative effect on the legibility is remedied by replacing the binarization with a greyscale transformation that smears out the transition from black to white around the threshold value (``soft thresholding'', see Fig.~\ref{fig:thresholding:c}).

Soft thresholding has some similarity with fuzzy thresholding \cite{sen09}, which assigns each grey value a ``membership value'' to foreground or background and sets pixels with a background membership value greater than 0.5 to white and the rest to black. Instead of utilizing the membership value for thresholding, it can be interpreted itself as a grey level, thereby defining a greyscale transformation. Fuzzy thresholding has two parameters, a membership function (typically Zadeh's ``S function'' \cite{zadeh78}) and a band width. Based on these, different criteria can be postulated to be optimized, thereby yielding a threshold value $t$. Even though there have been proposals for automatically determining the band width \cite{cheng97}, these do not work on all images and the band width must therefore in general be chosen manually to suit the grey-level histogram of the image \cite{murthy90}.

With fuzzy thresholding, the threshold is thus implicitly determined by the choice of the membership function and its band width and is not an independent variable. With soft thresholding, as presented in this report, the threshold is an independent variable that can be chosen to be optimal according to any other established method and the band width follows automatically from the threshold.

This report is organized as follows: in Sec.~\ref{sec:greyscaletransformations}, the term ``greyscale transformation'' is explained and appropriate transfer functions for soft thresholding are presented, in Sec.~\ref{sec:parameter}, a formula is given for computing the free parameter of these greyscale transformations for a given image, and Sec.~\ref{sec:local} gives examples how soft thresholding can be used with local thresholds.

A ready-to-run implementation of soft thresholding, as described in this report, has been implemented by the author within the free software {\em Gamera}\footnote{\url{http://gamera.sf.net/}}, a python library for building document analysis systems \cite{gamera}.

\section{Suitable greyscale transformations}
\label{sec:greyscaletransformations}
A {\em greyscale transformation} is a point operation on a greyscale image that replaces each pixel value $v$ by a new value $g(v)$ that depends only on the grey value of the pixel and not on its location or its neighborhood. Typical examples for greyscale transformations are contrast stretching and Gamma correction \cite{shi07}. A greyscale transformation is completely defined by its {\em transfer function} $g(v)$. Thresholding with threshold $t$ can also be considered as a greyscale transformation with the transfer function
\begin{equation}
g(v) = \left\{\begin{array}{ll} 0 & \mbox{ for }v\leq t \\
v_{\max} & \mbox{ for }v> t \end{array}\right.
\end{equation}
where $v_{max}$ is the highest grey value (255 for 8bit greyscale images). We can rewrite this as
\begin{subeqnarray}
\label{eq:transfer:thresholding}
\label{eq:transfer:thresholding:g}
g(v) & = & v_{max}\cdot F(v-t) \quad\mbox{ with} \\
\label{eq:transfer:thresholding:F}
F(z) & = & \left\{\begin{array}{ll} 0 & \mbox{ for }z\leq 0 \\
1 & \mbox{ for }z> 0 \end{array}\right.
\end{subeqnarray}
The ragged edges in Fig.~\ref{fig:thresholding:c} are due to the discontinuity of the function $F(z)$ at $z=0$. For soft thresholding, it is thus a natural generalization of Eq.~(\ref{eq:transfer:thresholding}) to replace $F(z)$ by a continuous nondecreasing function with the three properties
\begin{subeqnarray}
\label{eq:transfer:properties}
\lim_{z\to -\infty}F(z) & = & 0 \\
F(0) & = & 0.5 \\
\lim_{z\to\infty}F(z) & = & 1
\end{subeqnarray}

\begin{figure}[t]
  \begin{center}
    \includegraphics[width=\columnwidth]{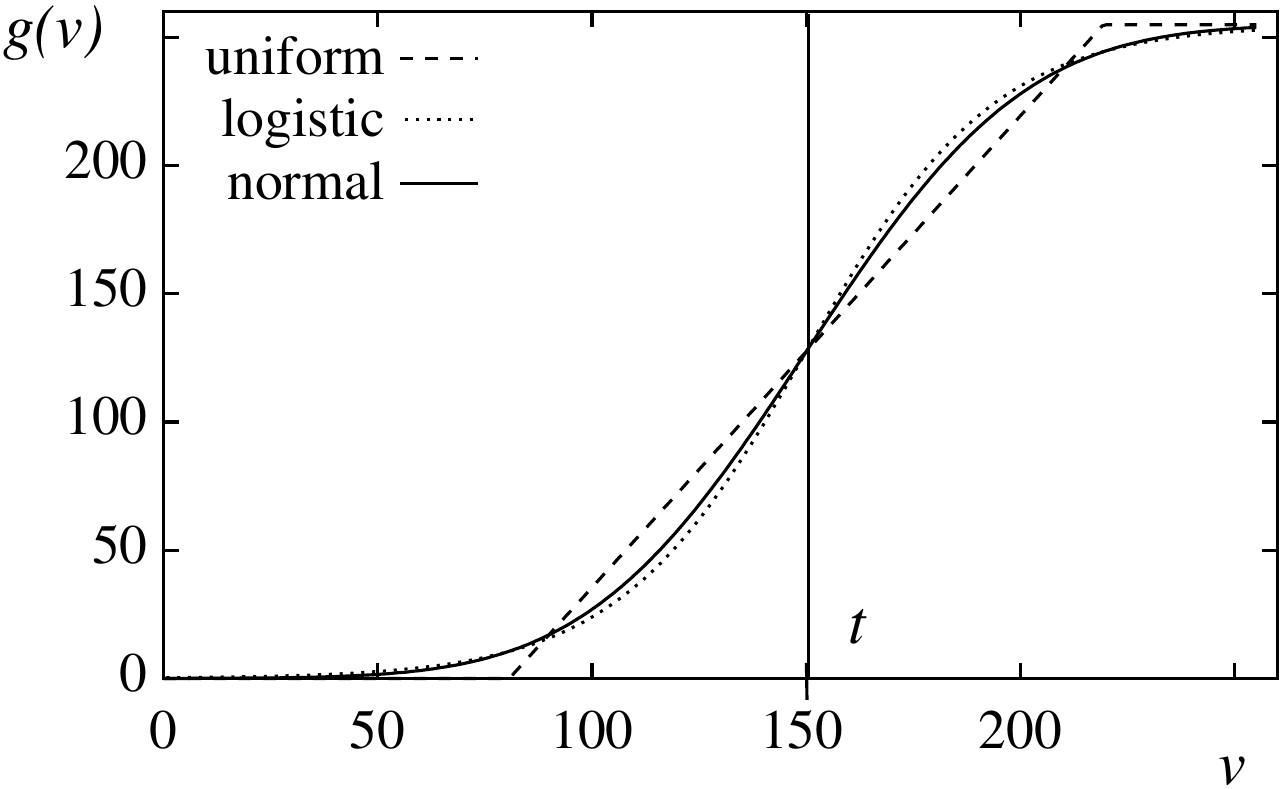}
  \end{center}
  \caption{\label{fig:cdfs}The transfer functions $g(v)$ defined in Eqs.~(\ref{eq:transfer:uniform}-\ref{eq:transfer:normal}). To make them comparable, the three functions are normalized to the same variance.}
\end{figure}

In other words, $F$ is the cumulative distribution function $F(z)=P(Z<z)$ of a probability distribution with median zero. The choice of the underlying probability distribution then determines the transfer function, e.g.
\begin{description}[font=\normalfont\itshape]
\item[Uniform distribution] \ 
\begin{equation}
\label{eq:transfer:uniform}
\hspace{-1.5em}g(v) = v_{max}\cdot \left\{\begin{array}{ll} 0 & \mbox{for }v\leq t-\frac{h}{2} \\[0.5ex]
\displaystyle \frac{v-t}{h} + \frac{1}{2} & \mbox{for } |v-t| < \frac{h}{2} \\[0.5ex]
1 & \mbox{for }v\geq t+\frac{h}{2} \end{array}\right.
\end{equation}
\item[Logistic or Fermi-Dirac distribution] \ 
\begin{equation}
\label{eq:transfer:logistic}
g(v) = \frac{v_{max}}{\displaystyle 1+\exp\left(-\frac{v-t}{\theta}\right)}
\end{equation}
\item[Normal distribution] \ 
\begin{equation}
\label{eq:transfer:normal}
g(v) = \frac{v_{max}}{2}\cdot\left(1-\mbox{erf}\left(\frac{v-t}{\sqrt{2\sigma^2}}\right)\right)
\end{equation}
\end{description}
where $h$ is the width of the uniform distribution, $\theta$ is the scale parameter (``temperature'') of the logistic distribution, $\sigma^2$ is the variance of the normal distribution, and {\em erf} is the error function $\mbox{erf}(z) = \frac{2}{\sqrt{\pi}}\int_0^z e^{-x^2}dx$.

The parameters $h$, $\theta$, and $\sigma$ are all proportional to the square root of the variance of the underlying probability distribution, and can thus be interpreted as a {\em band width}. As can be seen in Fig.~\ref{fig:cdfs}, these three transfer functions become somewhat similar when normalized to the same variance $\sigma^2$, i.e., with the choices
\begin{equation}
\label{eq:variance}
h=\sigma\cdot\sqrt{12} \quad\mbox{ and }\quad \theta=\sigma\cdot\frac{\sqrt{3}}{\pi}
\end{equation}
The transfer function based upon the normal distribution requires an implementation of the error function {\em erf}, which might not be available with all math libraries. As the transfer function based on the logistic distribution is quite similar, it can be used as a readily computable replacement.

\section{Parameter determination}
\label{sec:parameter}
The threshold $t$ in the transfer functions (\ref{eq:transfer:uniform}-\ref{eq:transfer:normal}) can be determined with any threshold selection algorithm, for example with Otsu's method \cite{otsu79}. The important question still remains how to choose the other parameter $\sigma$, $\theta$, or $h$ (note that these three parameters can be related through Eq.~(\ref{eq:variance})) in such a way that the result is visually an enhancement both compared to the original greyscale image and to the binary thresholded image.

Let $H(v)$ be the number of pixels in the image with grey value $v$. In other words, $H$ is the {\em grey-level histogram} of the image. As a threshold $t$ segments the pixels into the two classes ``black'' and ``white'', we can calculate the mean grey value $v_{w}$ in the ``white'' class as
\begin{equation}
\label{eq:meanwhite}
v_{w} = \frac{\sum_{v=t+1}^{v_{max}}v\cdot H(v)}{\sum_{v=t+1}^{v_{max}} H(v)}
\end{equation}
It is reasonable to expect from a ``soft thresholding'' algorithm that this value visually appears to be white, or is
\begin{equation}
\label{eq:conditionmeanwhite}
g(v_{w}) = \alpha\cdot v_{max} \quad\mbox{ with }\quad \alpha=0.99
\end{equation}

\begin{figure}[t]
  \begin{center}
    \includegraphics[width=\columnwidth]{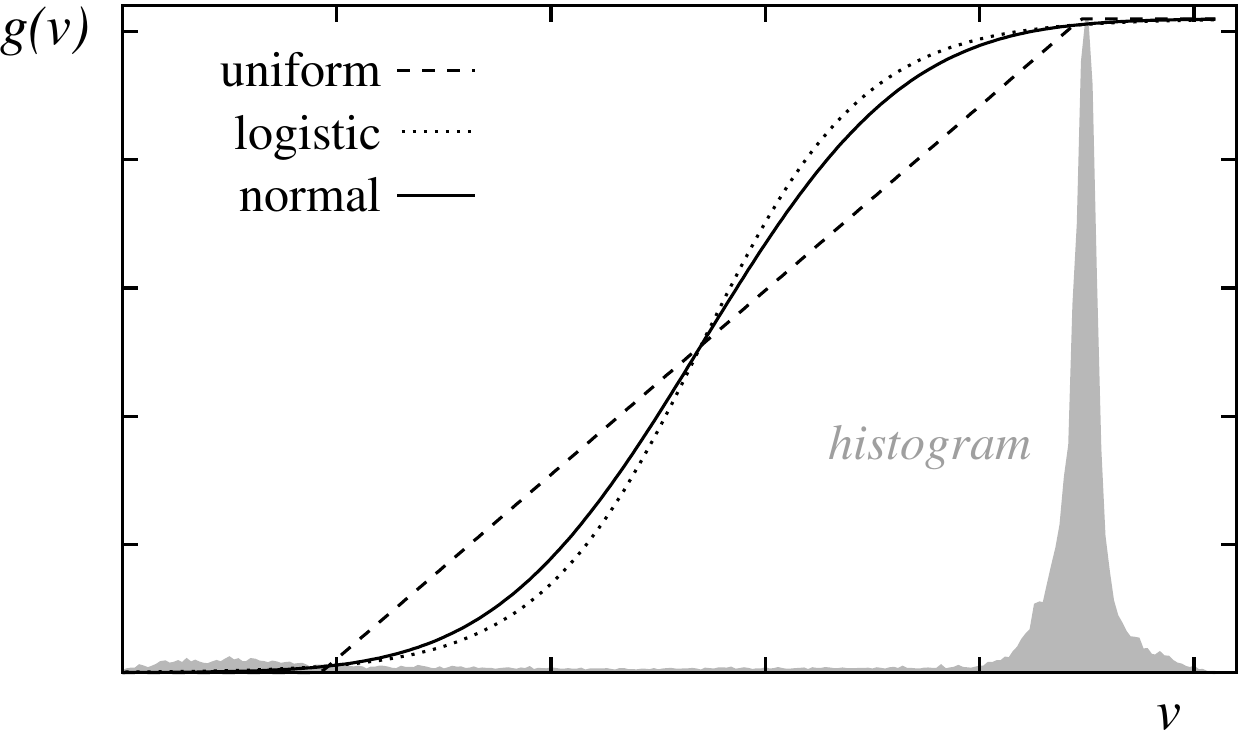}
  \end{center}
  \caption{\label{fig:histogram}A grey-level histogram (grey) and the transfer function resulting from Eqs.~(\ref{eq:param:uniform}-\ref{eq:param:normal}). Note that the histogram has been scaled to fill out the range $[0,v_{max}]$. The Otsu threshold for this histogram was $t=135$.}
\end{figure}

Substituting (\ref{eq:conditionmeanwhite}) into the three transfer functions (\ref{eq:transfer:uniform}-\ref{eq:transfer:normal}) and doing elementary calculations yields the following formulae for the parameter choice in the transfer functions:
\begin{description}[font=\normalfont\itshape]
\item[Uniform distribution] \ 
\begin{equation}
\label{eq:param:uniform}
h = \frac{v_w-t}{\alpha -0.5} \;\approx\; 2(v_w-t)
\end{equation}
\item[Logistic distribution] \ 
\begin{equation}
\label{eq:param:logistic}
\theta = -\frac{v_w-t}{\ln(-1+1/\alpha)}
\end{equation}
\item[Normal distribution] \ 
\begin{equation}
\label{eq:param:normal}
\sigma = \frac{v_w-t}{z_\alpha}
\end{equation}
\end{description}

\begin{figure*}[!t]
  \begin{center}
  \subfigure[Original]{
    \includegraphics[width=0.35\textwidth]{detail-grey}
    \label{fig:comparison:z}
  }\hspace{3em}
  \subfigure[Uniform]{
    \includegraphics[width=0.35\textwidth]{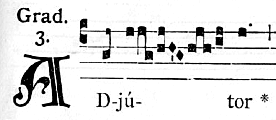}
    \label{fig:comparison:a}
  }\\
  \subfigure[Logistic]{
    \includegraphics[width=0.35\textwidth]{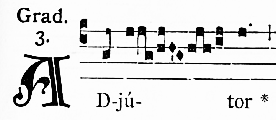}
    \label{fig:comparison:b}
  }\hspace{3em}
  \subfigure[Normal]{
    \includegraphics[width=0.35\textwidth]{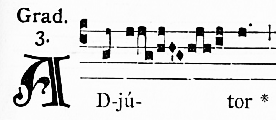}
    \label{fig:comparison:c}
  }
  \end{center}
  \caption{\label{fig:comparison}The effect of the distribution, upon which the transfer function is based, on the result of soft thresholding.}
\end{figure*}

where $z_\alpha$ is the $\alpha$-quantile of the standard normal distribution, which is $z_\alpha=2.3263$ for $\alpha=0.99$. Fig.~\ref{fig:histogram} shows the histogram of the image in Fig.~\ref{fig:comparison:z} and the resulting transfer functions based on the Otsu threshold and Eqs.~(\ref{eq:param:uniform}-\ref{eq:param:normal}). The effect of the different transfer functions can be seen in Fig.~\ref{fig:comparison}. The linear transition from black to white of the uniform distribution actually makes the slight shading at the left border more visible and does not suppress the show through from the back of the scanned page. The results for the normal and the logistic distributions are better and both quite similar, with the logistic distribution slightly better with respect to suppressing show through.

The function {\em soft\_threshold} in the Gamera framework therefore uses by default the transfer function based on the logistic distribution, and, when no threshold is provided by the user, Otsu's method is applied.

\section{Local thresholding}
\label{sec:local}
When the threshold $t=t(x,y)$ is not constant over the entire image, but depends on the pixel position $(x,y)$, it is no longer obvious how the parameters $h$, $\theta$, or $\sigma$ are to be determined. Depending on the local thresholding method, their values can be determined as follows:
\begin{enumerate}[label=\alph{enumi})]
\item When the local thresholding consists of a greyscale transformation followed by a global thresholding, the method from Sec.~\ref{sec:parameter} can be used.
\item When the thresholding algorithm computes the threshold $t(x,y)$ from the neighborhood of the pixel $(x,y)$, the same neighborhood can be used for computing a local $v_w(x,y)$ according to Eq.~(\ref{eq:meanwhite}), which can then be inserted into Eqs.~(\ref{eq:param:uniform}-\ref{eq:param:normal}) to obtain local parameters.
\item Global parameters can be obtained from a $v_w$ that is the mean grey value of all pixels assigned to class ``white'' by the thresholding algorithm.
\end{enumerate}

\lstset{language=Python,
  basicstyle=\small \ttfamily,
  frame=bottomline,
  floatplacement=!b,
  aboveskip=0pt,
  belowskip=0pt,
  captionpos=b
}
\begin{lstlisting}[float, caption=Python implementation for soft thresholding with shading subtraction utilizing the image processing functions provided by the Gamera framework., label=lst:shading]
# input  = grey image, filter size k
# output = soft thresholded image
def soft_shading_subtraction(image, k)
    shade = image.min_max_filter(k,1)
    shade = shade.to_float()
    imagef = image.to_float()
    diff = imagef.subtract_images(shade)
    diff = diff.to_greyscale()
    return diff.soft_threshold()
\end{lstlisting}

An example for a) is the shading subtraction described in \cite{toennies05}, which subtracts from each pixel value the maximum value of its $k\times k$ neighborhood and then performs a global thresholding on the resulting image. Note that the size $k$ must be chosen so large that a window always contains background pixels. As this cannot be guessed automatically, the filter size needs to be chosen manually by the user.

\begin{figure*}[t]
  \begin{center}
  \subfigure[Greyscale image]{
    \includegraphics[width=0.75\textwidth]{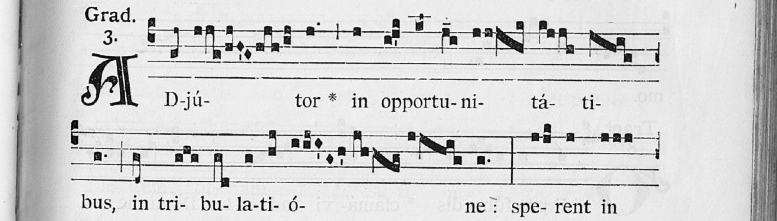}
    \label{fig:shading:a}
  }
  \subfigure[Global soft thresholding]{
    \includegraphics[width=0.75\textwidth]{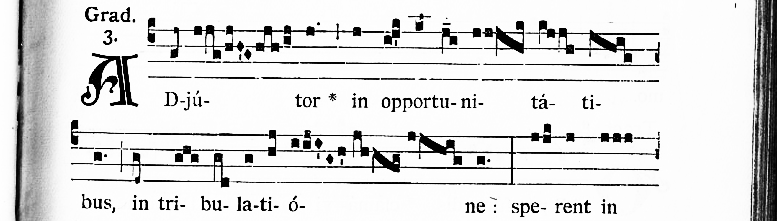}
    \label{fig:shading:b}
  }
  \subfigure[Soft thresholding after shading subtraction]{
    \includegraphics[width=0.75\textwidth]{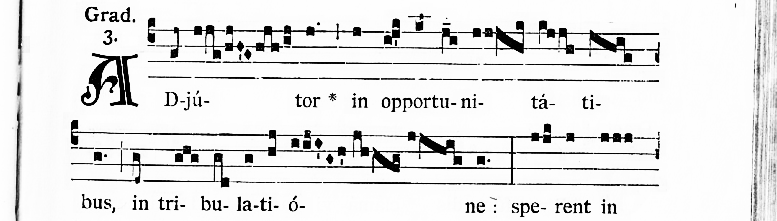}
   \label{fig:shading:c}
 }
  \end{center}
  \caption{\label{fig:shading}A greyscale image with shading on the left edge globally soft thresholded with Otsu's threshold and after shading subtraction ($k=17$) according to Listing \ref{lst:shading} (image detail from ``Graduale Romanum'', Tournai, 1910).}
\end{figure*}

As a binarization method, this local thresholding is implemented in the Gamera function {\em shading\_subtraction} with the use of a fast maximum filter implementation based on Ref.~\cite{gil93}. An adaption of this method for soft thresholding is given in Listing \ref{lst:shading}, and the result can be seen in Fig.~\ref{fig:shading}. Compared to Fig.~\ref{fig:shading:b}, the shadow from the book binding is absent in Fig.~\ref{fig:shading:c}.

\section{Conclusions}
\label{sec:conclusions}
The soft thresholding algorithm presented in this report is a greyscale transformation that can be used to visually enhance scanned document images. When the scans have regions with varying illumination (shading), as typically occurs with thick books due to the book binding, the combination of soft thresholding with shading subtraction yields decent results.

The author has made a freely available implementation of soft thresholding within the Gamera framework for document analysis and recognition. As this is a python library function, and as it determines the parameters for soft thresholding automatically when no parameters are provided by the user, the method is usable out-of-the-box to automatically process large repositories of online facsimiles. It is also useful as a superior alternative to binarization for preparing images for printed facsimile editions.

\bibliographystyle{ieeetr}
\bibliography{soft-thresholding}

\end{document}